\documentclass[11pt]{article}

\usepackage[]{acl}

\usepackage{times}
\usepackage{latexsym}

\usepackage{amsfonts}
\usepackage[T1]{fontenc}

\usepackage[utf8]{inputenc}

\usepackage{microtype}

\usepackage{inconsolata}

\usepackage{graphicx}
\usepackage{amsmath}
\usepackage{booktabs}
\usepackage{multirow}
\usepackage{makecell}
\usepackage{amssymb}
\usepackage{txfonts}
\usepackage[table]{xcolor}

\title{Dynamic Meta-Metrics: Source-Sentence Conditioned Weighting for MT Evaluation}

\author{
 \textbf{Luke Zhang$^{\varheartsuit}$},\
 \textbf{Justin Vasselli$^{\clubsuit}$},\
 \textbf{Aditya Khan$^{\varheartsuit}$},\
 \textbf{York Hay Ng$^{\varheartsuit}$},\
 \textbf{En-Shiun Annie Lee$^{\varheartsuit\hspace{1pt}\vardiamondsuit}$}
\\
 $^{\varheartsuit}$University of Toronto, Canada\quad
 $^{\clubsuit}$Nara Institute of Science and Technology, Japan\\
 $^{\vardiamondsuit}$Ontario Tech University, Canada
 \\
    \texttt{
        lukelz.zhang@mail.utoronto.ca, 
        vasselli\_justinray.vk4@is.naist.jp
        } \\
    \texttt{
        adityakhan@cs.toronto.edu,
        yorkng@cs.toronto.edu  
 }
}

\newif\ifshowcomments
\showcommentstrue 

\begin{document}
\maketitle
\begin{abstract}
We propose Dynamic Meta-Metrics (DMM), a framework for machine translation evaluation that learns source-sentence conditioned combinations of existing metrics. Rather than relying on a single static ensemble or language-specific weighting, DMM adapts the metric combination based on properties of the source segment. We study hard conditioning, which fits an interpretable combiner per cluster, and an exploratory soft-conditioned extension whose weights vary continuously with source-cluster responsibilities. We evaluate DMM on the WMT Metrics Shared Task data across multiple language pairs using pairwise agreement measures at the system and segment levels. Across settings, MLP-based combinations outperform linear and Gaussian process-based ensembles, and introducing soft conditioning yields gains over linear models.

\end{abstract}

\section{Introduction}

Automatic evaluation metrics underpin machine translation (MT) research and deployment. An effective metric should align with human judgements, generalise across languages and domains, and remain reproducible. The WMT Metrics Shared Task provides a standardised setting for meta-evaluation and has shown that metric behaviour varies with language, domain, and input characteristics \citep{freitag-etal-2022-results,kocmi-etal-2023-findings,freitag-etal-2024-llms}. In particular, no single metric consistently dominates across years and test conditions, even among strong neural and generative metrics \citep{juraska-etal-2023-metricx,juraska-etal-2024-metricx,freitag-etal-2024-llms}.

\citet{anugraha-etal-2024-metametrics}, following \citet{winata_metametrics}, addressed distribution shift by combining multiple metrics into static ensembles with fixed weights, which can be more robust than relying on a single metric, termed \emph{meta-metrics} (see Appendix \ref{app:related-work}). This is not to be confused with meta-metrics in the context of \emph{evaluating} metrics (e.g. as in \citet{thompson-etal-2024-improving}). However, such ensembles do not model variation within a language pair, where differences in syntax, discourse style, or domain can affect metric reliability.

\begin{figure}[t]
    \centering
    \includegraphics[width=1.0\linewidth]{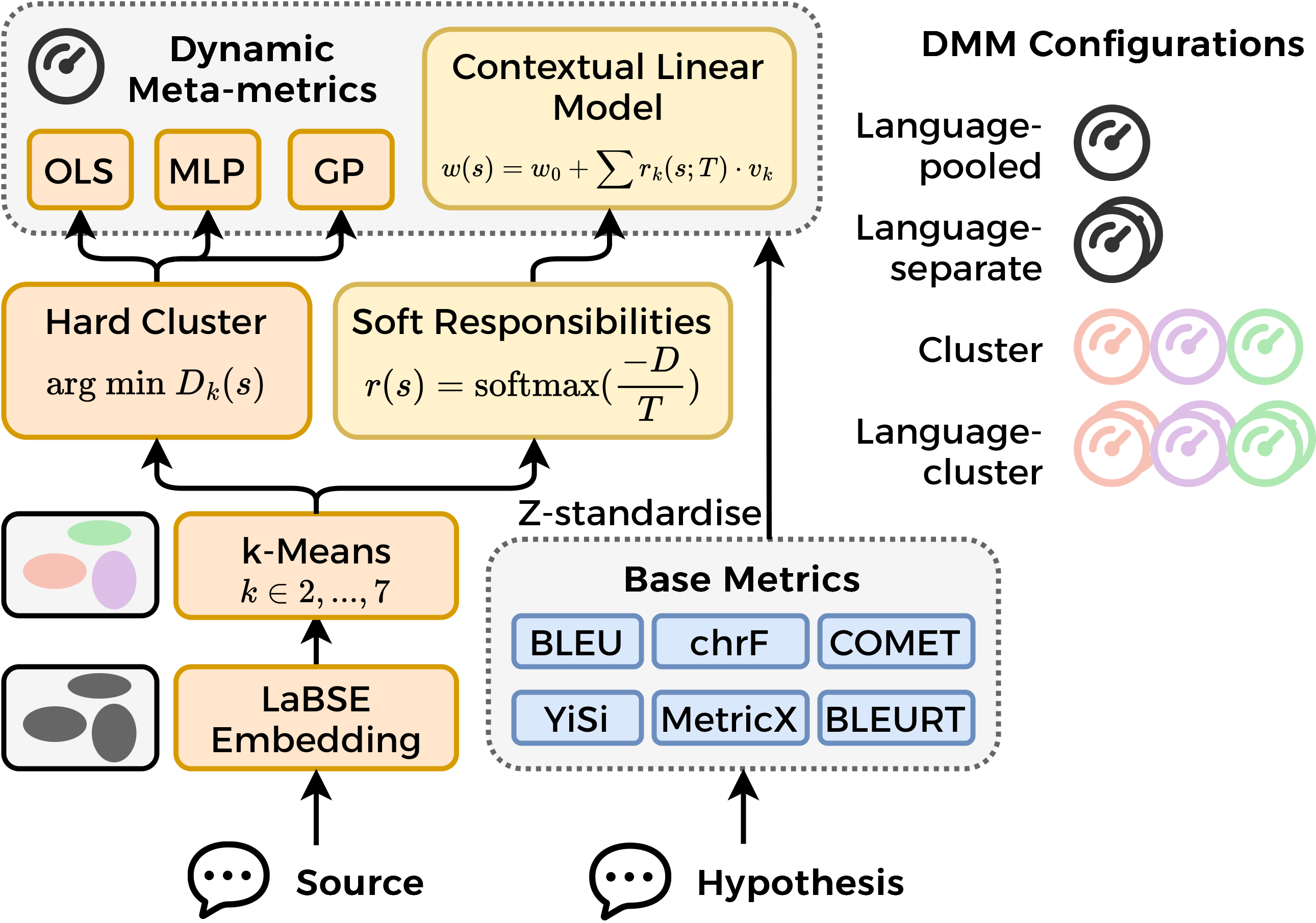}
    \caption{The DMM framework combines MT evaluation metrics conditioned on source-sentence context, with four training configurations.}
    \label{fig:placeholder}
\end{figure}

To address this limitation, we introduce \emph{Dynamic Meta-Metrics} (DMM), a framework for MT evaluation that conditions metric combination on source-sentence context rather than using a single fixed ensemble. DMM constructs a discrete context variable from sentence embeddings and clustering: each source sentence is embedded with LaBSE~\citep{feng-etal-2022-language} and grouped using $k$-means, after which the model learns context-specific combinations of base metric scores. We call this \textbf{hard conditioning}. This preserves interpretability by yielding explicit weights for each context while allowing metric contributions to vary with the input. We instantiate DMM with three meta-metric models: (i) a linear regressor (fit with ordinary least squares -- OLS) minimising MSE on segment-level human scores, (ii) a small multilayer perceptron (MLP) trained with the mean-squared error loss (MSE), and (iii) a Gaussian-process-driven Bayesian optimiser that maximises Kendall correlation to human scores.

Hard conditioning assigns each source segment to a single cluster and applies the corresponding combiner. This yields an interpretable piecewise-constant rule, but it may be restrictive when a source segment exhibits properties shared by several source contexts. We therefore also define an exploratory \textbf{soft-conditioned} extension of DMM, which uses the source segment's distances to all cluster centroids to form a \emph{soft} contextual representation, allowing the final metric weights to vary continuously with context.

Our contributions are threefold: (i) we propose DMM, a source-conditioned framework for combining MT evaluation metrics; (ii) we introduce hard and soft conditioning variants that trade off interpretability and flexibility; and (iii) we show on WMT data \citep{freitag-etal-2024-llms} across four representative language pairs that MLP-based combinations outperform linear and Gaussian process ensembles, and that soft conditioning yields substantial gains over linear models.

\section{Method}\label{sec:method}
\subsection{Problem setup and notation}
Each training instance is a triple $(s, t, y)$, where $s$ denotes a source segment, $t$ a system output (hypothesis) for that segment, and $y \in \mathbb{R}$ a human segment-level score. We index instances by $i \in \{1,\ldots,n\}$. For each instance $i$, we compute $d$ base metric scores and form a feature vector
$\mathbf{x}_i = \big(m_j(i)\big)_{j=1}^d \in \mathbb{R}^d$, with $m_j(i)$ being the $j$th metric score for instance $i$.
A meta-metric is a function $F_\theta : \mathbb{R}^d \rightarrow \mathbb{R}$ that predicts $\hat{y}_i = F_\theta(\mathbf{x}_i)$, with learnable parameters $\theta$.

\subsection{Feature standardisation}
Since metric outputs differ in scale, we $z$-standardise each metric feature using training-set statistics. For metric $j$, let $\mu_j$ and $\sigma_j$ be the mean and standard deviation over training instances. We standardise $x_{ik}$, obtaining $\tilde x_{ik}$ forming $\tilde{\mathbf{x}}_i = (\tilde{x}_{i1}, \ldots, \tilde{x}_{id})$. All subsequent linear and neural models use $\tilde{\mathbf{x}}_i$. Note that standardisation does not require the metric-score distribution to be Gaussian. It is only to put the scores on a common scale.

\subsection{Embedding and clustering source segments}
We embed each source segment $s$ using LaBSE \citep{feng-etal-2022-language}, yielding $\mathbf{e}(s) \in \mathbb{R}^m$. The reason we use LaBSE is because DMM requires a source-only representation that is comparable across language pairs. LaBSE is designed for
language-agnostic sentence embeddings and provides a fixed multilingual sentence space, making it a natural
choice for clustering source segments across languages.

Following that, we fit $k$-means on the training-set source embeddings to obtain $K$ centroids $\{\mathbf{c}_1,\ldots,\mathbf{c}_K\}$. For any segment $s$, we define squared distances $D_k(s) = \lVert\mathbf{e}(s) - \mathbf{c}_k\rVert_2^2$ for $k=1, ..., K$.

\subsection{Soft-conditioned DMM}\label{subsec:soft-dmm}
We now turn to the soft-conditioning extension. For a source segment $s$, the squared distances from its embedding to the $K$ cluster centroids are converted into a soft responsibility vector $\mathbf{r}(s;T)=\text{softmax}(\{-D_k(s)/T\}_{k =1}^K)$, where $T>0$ is a temperature parameter. Smaller values of $T$ produce sharper, near one-hot assignments, while larger values yield more diffuse mixtures over clusters. The responsibilities depend only on the source segment and are therefore shared across all candidate translations of that segment.

We then define a contextual linear meta-metric whose effective weight vector varies with source context:
$\mathbf{w}_{\mathrm{eff}}(s;T) = \mathbf{w}_0+\sum_{k=1}^{K} r_k(s;T)\,\mathbf{v}_k,$
where $\mathbf{w}_0\in\mathbb{R}^d$ is a global baseline weight vector and $\mathbf{v}_1,\ldots,\mathbf{v}_K\in\mathbb{R}^d$ are cluster-associated deviation vectors. The prediction for instance $i$ is
$\hat y_i=\tilde{\mathbf{x}}_i^\top \mathbf{w}_{\mathrm{eff}}(s_i;T)+b$
with intercept $b\in\mathbb{R}$.

This constitutes \emph{soft conditioning} because each source segment induces a continuous weighting over clusters, so the metric-combination weights vary smoothly with that weighting. Further details such as how 1) hard conditioning appears as a special case of this model, and 2) why we do not use a neural model for soft conditioning are described in Appendix \ref{app:soft-dmm}.

\subsection{Ensemble Configuration}
\label{sec:hard_conditioned}
We consider two orthogonal design choices: (i) whether training data is pooled across language pairs or separated by language, and (ii) whether clustering is used. In particular, the configurations are as follows.

\begin{itemize}
    \item \textbf{Language-pooled:} a single model is trained on all instances across language pairs.
    \item \textbf{Language-separate:} one model is trained per language pair.
    \item \textbf{MM (meta-metrics; no conditioning):} a single model is applied within each training setup. This corresponds to a standard meta-metric.
    \item \textbf{DMM (dynamic meta-metrics;  conditioning):} data is partitioned into $K$ clusters based on the source segment, and a separate model is trained for each cluster.
\end{itemize}

For each configuration, we train multiple model classes: ordinary least squares (OLS), Gaussian processes (GP), and multilayer perceptrons (MLP). For DMM, we additionally include a contextual linear (CL) variant described in Section~\ref{subsec:soft-dmm}.


\subsection{Evaluation measures}
Following WMT24~\citep{freitag-etal-2024-llms}, we evaluate metrics primarily using pairwise agreement with human judgements. At the system level, we report soft pairwise accuracy (SPA)~\citep{thompson-etal-2024-improving}, which accounts for uncertainty in both human and metric-induced rankings. At the segment level, we report group-by-item pairwise accuracy with tie calibration ($acc^*_{eq}$)~\citep{deutsch-etal-2023-ties}, reflecting the metric’s ability to rank alternative translations of the same source segment correctly.

\begin{table*}[!t]
  \centering
  \small
  \setlength{\tabcolsep}{5pt}
  \renewcommand{\arraystretch}{1.12}
  \begin{tabular}{l cc cc cc cc cc}
    \toprule
    \multirow{2}{*}{\textbf{System}}
      & \multicolumn{2}{c}{\textbf{EN-CS}}
      & \multicolumn{2}{c}{\textbf{EN-ZH}}
      & \multicolumn{2}{c}{\textbf{EN-JA}}
      & \multicolumn{2}{c}{\textbf{EN-UK}}
      & \multicolumn{2}{c}{\textbf{Avg}} \\
    \cmidrule(lr){2-3}\cmidrule(lr){4-5}\cmidrule(lr){6-7}\cmidrule(lr){8-9}\cmidrule(lr){10-11}
    & $\boldsymbol{acc^*_{eq}}$ & \textbf{SPA}
    & $\boldsymbol{acc^*_{eq}}$ & \textbf{SPA}
    & $\boldsymbol{acc^*_{eq}}$ & \textbf{SPA}
    & $\boldsymbol{acc^*_{eq}}$ & \textbf{SPA}
    & $\boldsymbol{acc^*_{eq}}$ & \textbf{SPA} \\
    \midrule
    Gemini-2.5-Pro & \cellcolor{blue!29}0.613 & \cellcolor{blue!26}0.890 & \cellcolor{blue!8}0.530 & \cellcolor{blue!30}\textbf{0.851} & \cellcolor{blue!20}0.559 & \cellcolor{blue!30}\textbf{0.825} & \cellcolor{blue!4}0.517 & \cellcolor{blue!30}\textbf{0.819} & \cellcolor{blue!18}0.559 & \cellcolor{blue!30}\textbf{0.851} \\
    wmt22-comet-da & \cellcolor{blue!18}0.595 & 0.685 & \cellcolor{blue!14}0.544 & \cellcolor{blue!1}0.626 & \cellcolor{blue!9}0.536 & \cellcolor{red!5}0.526 & \cellcolor{blue!30}\textbf{0.572} & \cellcolor{blue!4}0.696 & \cellcolor{blue!20}0.562 & 0.633 \\
    gemba & \cellcolor{red!30}0.369 & \cellcolor{blue!10}0.814 & \cellcolor{red!30}0.353 & \cellcolor{blue!17}0.793 & \cellcolor{red!30}0.341 & \cellcolor{blue!3}0.725 & \cellcolor{red!30}0.346 & \cellcolor{blue!10}0.741 & \cellcolor{red!30}0.352 & \cellcolor{blue!10}0.768 \\
    \midrule
    \multicolumn{11}{l}{\textsc{Language-Pooled MM}} \\
    \quad GP & \cellcolor{blue!25}0.607 & \cellcolor{blue!24}0.883 & \cellcolor{blue!16}0.548 & \cellcolor{blue!3}0.666 & \cellcolor{blue!28}0.570 & \cellcolor{blue!8}0.757 & \cellcolor{blue!12}0.543 & \cellcolor{blue!6}0.718 & \cellcolor{blue!23}0.567 & \cellcolor{blue!8}0.756 \\
    \quad OLS & 0.486 & \cellcolor{red!18}0.380 & 0.473 & \cellcolor{red!10}0.276 & \cellcolor{blue!1}0.494 & \cellcolor{red!20}0.465 & 0.472 & \cellcolor{red!14}0.372 & 0.481 & \cellcolor{red!15}0.373 \\
    \quad MLP & \cellcolor{blue!29}0.612 & \cellcolor{blue!29}0.898 & \cellcolor{blue!30}\textbf{0.568} & \cellcolor{blue!22}0.818 & \cellcolor{blue!25}0.566 & \cellcolor{blue!5}0.741 & \cellcolor{blue!19}0.556 & \cellcolor{blue!4}0.692 & \cellcolor{blue!30}\textbf{0.576} & \cellcolor{blue!13}0.787 \\
    \midrule
    \multicolumn{11}{l}{\textsc{Language-Separate MM}} \\
    \quad GP & \cellcolor{blue!11}0.578 & \cellcolor{blue!2}0.728 & \cellcolor{blue!20}0.554 & \cellcolor{blue!1}0.608 & \cellcolor{blue!20}0.559 & \cellcolor{blue!10}0.766 & \cellcolor{blue!15}0.548 & \cellcolor{blue!12}0.752 & \cellcolor{blue!19}0.560 & \cellcolor{blue!3}0.714 \\
    \quad OLS & 0.494 & \cellcolor{red!16}0.390 & \cellcolor{blue!3}0.510 & 0.467 & \cellcolor{blue!9}0.534 & 0.640 & \cellcolor{blue!1}0.488 & \cellcolor{red!4}0.440 & \cellcolor{blue!2}0.507 & \cellcolor{red!2}0.485 \\
    \quad MLP & \cellcolor{blue!30}\textbf{0.614} & \cellcolor{blue!30}\textbf{0.902} & \cellcolor{blue!29}0.567 & \cellcolor{blue!15}0.782 & \cellcolor{blue!26}0.567 & \cellcolor{blue!5}0.739 & \cellcolor{blue!20}0.557 & \cellcolor{blue!6}0.715 & \cellcolor{blue!30}\textbf{0.576} & \cellcolor{blue!13}0.784 \\
    \midrule
    \multicolumn{11}{l}{\textsc{Language-Pooled DMM (Clusters)} ($k=6$)} \\
    \quad GP & \cellcolor{blue!11}0.578 & \cellcolor{blue!6}0.781 & \cellcolor{blue!19}0.553 & \cellcolor{blue!1}0.633 & \cellcolor{blue!25}0.566 & \cellcolor{blue!4}0.728 & \cellcolor{blue!18}0.555 & \cellcolor{blue!2}0.664 & \cellcolor{blue!21}0.563 & \cellcolor{blue!3}0.702 \\
    \quad OLS & 0.470 & \cellcolor{red!27}0.345 & 0.463 & \cellcolor{red!26}0.190 & 0.479 & \cellcolor{red!30}0.441 & 0.464 & \cellcolor{red!16}0.364 & 0.469 & \cellcolor{red!25}0.335 \\
    \quad MLP & \cellcolor{blue!27}0.610 & \cellcolor{blue!25}0.887 & \cellcolor{blue!29}0.567 & \cellcolor{blue!21}0.815 & \cellcolor{blue!22}0.562 & \cellcolor{blue!7}0.749 & \cellcolor{blue!18}0.554 & \cellcolor{blue!3}0.679 & \cellcolor{blue!28}0.573 & \cellcolor{blue!12}0.782 \\
    \quad CL & \cellcolor{blue!16}0.591 & \cellcolor{blue!1}0.709 & \cellcolor{blue!13}0.542 & 0.599 & \cellcolor{blue!30}\textbf{0.573} & \cellcolor{blue!4}0.735 & \cellcolor{blue!14}0.547 & \cellcolor{blue!1}0.639 & \cellcolor{blue!21}0.563 & \cellcolor{blue!1}0.670 \\
    \midrule
    \multicolumn{11}{l}{\textsc{Language-Separate DMM (Clusters)} ($k=6$)} \\
    \quad GP & \cellcolor{blue!8}0.570 & \cellcolor{blue!1}0.702 & \cellcolor{blue!8}0.531 & \cellcolor{blue!3}0.673 & \cellcolor{blue!16}0.551 & \cellcolor{blue!11}0.770 & \cellcolor{blue!16}0.551 & \cellcolor{blue!11}0.748 & \cellcolor{blue!14}0.551 & \cellcolor{blue!4}0.723 \\
    \quad OLS & 0.464 & \cellcolor{red!30}0.334 & 0.449 & \cellcolor{red!30}0.175 & \cellcolor{blue!1}0.494 & \cellcolor{red!24}0.454 & 0.467 & \cellcolor{red!30}0.315 & 0.469 & \cellcolor{red!30}0.320 \\
    \quad MLP & \cellcolor{blue!29}0.613 & \cellcolor{blue!28}0.896 & \cellcolor{blue!29}0.567 & \cellcolor{blue!20}0.807 & \cellcolor{blue!26}0.568 & \cellcolor{blue!6}0.743 & \cellcolor{blue!18}0.555 & \cellcolor{blue!3}0.682 & \cellcolor{blue!30}\textbf{0.576} & \cellcolor{blue!12}0.782 \\
    \quad CL & \cellcolor{blue!13}0.585 & \cellcolor{blue!1}0.700 & \cellcolor{blue!11}0.537 & 0.574 & \cellcolor{blue!28}0.571 & \cellcolor{blue!2}0.716 & \cellcolor{blue!12}0.543 & \cellcolor{blue!1}0.634 & \cellcolor{blue!18}0.559 & \cellcolor{blue!1}0.656 \\
    \bottomrule
  \end{tabular}
  \caption{Segment- and system-level meta-evaluation results. We report group-by-item pairwise accuracy with tie calibration ($acc^*_{eq}$) and soft pairwise accuracy (SPA), comparing our MM and DMM variants against Gemini-2.5-Pro-as-a-Judge, the top system at WMT25, as well as wmt22-comet-da and gemba, the top submetrics used by average $acc^*_{eq}$ and SPA respectively. Higher is better.}
  \label{tab:detailed_results}
\end{table*}

\section{Experiments}
\subsection{Experimental Setup}

We use WMT Metrics Shared Task data from 2021--2024 for training and validation
\citep{freitag-etal-2021-results,freitag-etal-2022-results,freitag-etal-2023-results,freitag-etal-2024-llms},
including English to Czech (en-cs), Chinese (en-zh), Ukrainian (en-uk), and Japanese (en-ja).
We evaluate on held-out WMT25 data, citing both the WMT25 Automated Translation Evaluation shared task,
which defines the metric-evaluation setting, and the WMT25 General MT shared task, which provides the
underlying translation outputs and human evaluation context
\citep{lavie-etal-2025-findings,kocmi-etal-2025-findings}.
For each pair and year, we use source segments, system outputs, reference translations, and official
segment-level human annotations, excluding segments without human scores. We use a fixed, deterministic split defined at the segment level within each year (approximately 80\% train, 20\% validation), created independently per year. We evaluate on held-out WMT25 shared task data.

We compute a representative set of reference-based metrics spanning overlap, embedding-based, and generative families, using default toolkit settings unless stated otherwise. We exclude the largest metric variants (e.g., MetricX-XXL) to support execution on typical hardware. We consider SacreBLEU, BLEU, and sentence-level BLEU variants \citep{papineni-etal-2002-bleu,post-2018-call}, chrF++ \citep{popovic-2017-chrf}, BLEURT-20 \citep{sellam-etal-2020-bleurt,pu-etal-2021-learning}, two variants of YiSi-1 \citep{lo-2019-yisi} (using XMLRoberta \citep{conneau-etal-2020-unsupervised} and LaBSE \citep{feng-etal-2022-language}), COMET, and XCOMET-XL \citep{rei-etal-2020-comet}, MetricX-23/24 Large and XL \citep{juraska-etal-2023-metricx,juraska-etal-2024-metricx}, and GEMBA-MQM \citep{kocmi-federmann-2023-gemba} using Qwen3-30B \citep{yang2025qwen3technicalreport} as the underlying model.

For model fitting, we embed source segments with LaBSE and fit $k$-means on training sources. We evaluate $K \in \{2,3,4,5,6,7\}$. Implementation details, hyperparameters, and model selection criteria are provided in Appendix \ref{app:implementation}.

\subsection{Results}

Table~\ref{tab:detailed_results} reports system- and segment-level meta-evaluation results across language pairs, comparing baseline metrics with the proposed DMM variants under different conditioning strategies.

\begin{table*}[!t]
\centering
\footnotesize
\setlength{\tabcolsep}{3pt}
\renewcommand{\arraystretch}{1}
\begin{tabular}{l r r p{3.3cm} p{4.8cm} p{4cm}}
\toprule
\textbf{Cluster}
  & \textbf{$|S|$}
  & \textbf{\makecell{Med. \\ Tok}}
  & \textbf{Domains}
  & \textbf{GP Top}
  & \textbf{MLP Top} \\
\midrule
0 & 1,393 & 17 & Minutes, e-comm., social    & MetricX-24-L, XCOMET & spBLEU, chrF \\
1 & \phantom{0}871  & 11 & Social, news, e-comm.       & BLEU, MetricX-24-L,MetricX-24-XL & MetricX-24-L, MetricX-24-XL \\
2 & \phantom{0}998  & 25 & News (NYT, BBC)          & MetricX-24-L            & MetricX-24-L \\
3 & \phantom{0}789  & 24 & News, literary     & MetricX-24-L,MetricX-24-XL       & MetricX-24-L \\
4 & 1,482 & \phantom{0}7 & E-comm., social, minutes    & BLEU      & BLEU \\
5 & \phantom{0}963  & 66 & Literary (cross-lingual)     & COMET                   & COMET \\
\bottomrule
\end{tabular}
\caption{
  $|S|$: segments; Med.\ Tok: median token count;
  If top and second metric are within 0.01 then both are listed.
}
\label{tab:cluster_metrics_wide}
\end{table*}

\paragraph{Performance.} Table~2 reports system- and segment-level results. For DMM, we report $K=6$, which performed best on the validation set (having tried $K=2, ..., 7$). The MLP combiner achieves the highest average performance across all configurations. All MLP variants outperform Gemini-2.5-Pro at the segment level (avg $acc^*_{eq}$ 0.576 vs.\ 0.559), while Gemini retains a system-level SPA advantage (0.851 vs.\ 0.787). Cluster-conditioned DMM performs comparably to static MM for the MLP combiner: the best DMM configuration (language-separate, $K=6$) achieves avg $acc^*_{eq}$ 0.576 / SPA 0.782 vs.\ 0.576 / 0.784 for language-separate MM, suggesting that MLP's nonlinear hidden layers already capture source-context-dependent behaviour that clustering makes explicit for simpler models. The GP combiner shows greater sensitivity to conditioning but lower absolute performance, and OLS generalises poorly across all settings. The soft-conditioned contextual linear model (CL) underperforms the hard-conditioned variants (avg SPA 0.670 for language-pooled, $K=6$), which may reflect overfitting in the expanded parameter space.

\paragraph{Cluster interpretability.}
We profile the $K=6$ clusters to examine what source-level structure DMM exploits. Clusters separate primarily by sentence length and domain rather than language pair (Table~1). For example, Cluster~5 contains longer literary segments (median 66 tokens), Clusters~1 and 4 contain short segments (7--11 tokens) from e-commerce and social media, and Clusters~2--3 contain medium-length news text (24--25 tokens). 

We next examine per-cluster metric preferences using GP weights and MLP feature attributions. In semantically coherent clusters, both models agree on the dominant metric: BLEU for short segments and MetricX-24-Large for news, while longer literary segments favor neural metrics such as COMET. Agreement is weaker in mixed or transitional clusters, where surface and neural metrics provide comparable signal.
Overall, these patterns suggest that clustering captures meaningful variation in metric reliability, with metric preferences aligning with known strengths of lexical and neural metrics.

\begin{figure}[!t]
  \includegraphics[width=\linewidth]
  {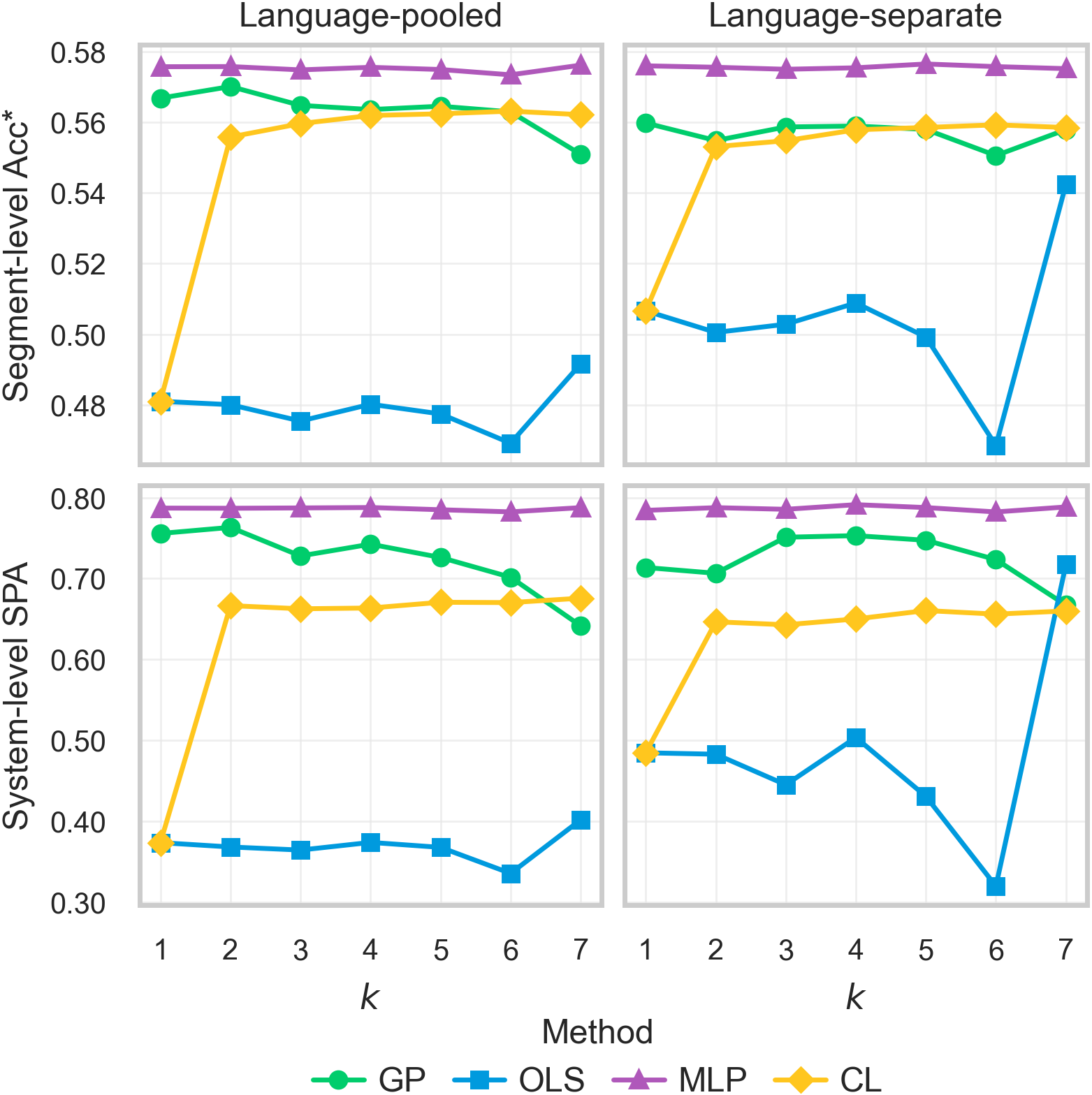}
  \caption {Effect of the number of clusters $k$ on meta-metric performance. Top panels show Acc*, bottom panels show SPA across language pairs.}
  \label{fig:clusters}
\end{figure}

\paragraph{Effect of $k$.}
Figure \ref{fig:clusters} shows the effect of the number of clusters $k$ on performance. Across settings, performance is largely insensitive to $k$. The MLP model remains stable across all values of $k$, suggesting it captures source-dependent structure without explicit clustering. In contrast, GP and OLS models are more sensitive to $k$, with performance degrading under finer-grained clustering. Notably, the contextual linear (CL) model benefits the most from clustering, while remaining stable across $k\in[2,7]$.
\paragraph{Cross-lingual transfer.} Language-pooled configurations perform comparably to language-separate ones across all combiners (Table~2: MLP avg $acc^*_{eq}$ 0.576 / SPA 0.787 pooled vs.\ 0.576 / 0.784 separate for static MM; similar patterns hold under clustering). This provides evidence that source-sentence semantics, captured by LaBSE embeddings, are more predictive of metric reliability than language identity, and that combination strategies learned on pooled multilingual data transfer across language pairs.

\paragraph{Unseen language pairs.}
Table \ref{tab:unseen_language_pairs} reports results on language pairs not observed during training. Overall, the MLP-based meta-metric remains the strongest performer, showing stable generalisation across languages. In contrast, DMM with hard clustering does not provide consistent gains over standard meta-metrics in this setting. As in the seen-language experiments, Gaussian process and linear models degrade under hard clustering. Notably, the contextual linear (CL) model substantially improves over static linear regression, indicating that soft conditioning provides a more robust mechanism for adapting metric weights under distribution shift.

\section{Conclusion}

We introduced Dynamic Meta-Metrics (DMM), a framework for combining MT evaluation metrics based on the properties of the source segment.
Across experiments, model choice and conditioning play a larger role than clustering itself. MLP-based combinations consistently outperform linear and GP-based ensembles and remain stable across seen and unseen language pairs. In contrast, hard clustering provides limited benefit and can degrade performance, particularly under distribution shift, while soft conditioning substantially improves linear models.
These results suggest that source-dependent weighting is beneficial, but that flexible conditioning mechanisms are more robust than discrete clustering for meta-metric design.

\section*{Limitations}
\paragraph{Source-only conditioning.}
Both hard and soft conditioning derive their context from the source segment. This is to make the routing system-invariant. However, this still does not directly capture any other phenomena, such as hypothesis-specific difficulty. 

\paragraph{Metric set constraints.}
We exclude the largest metric variants to support execution on typical hardware. This may limit direct comparison with the strongest available single-metric baselines.

\paragraph{Metric-on-metric overfitting.} When optimising for MSE (Linear Regressors) or correlation (Gaussian Processes), the combiner only sees other metrics’ scores instead of human error patterns, so it can chase artefacts of those metrics (e.g., overlap bias, domain-specific scaling). For example, if a base metric has a quirk (e.g., it over-rewards literal overlap on speech transcripts), the weight search can “learn” that quirk as a shortcut to higher validation scores without actually getting closer to human judgement.

\paragraph{Language coverage.} We train on only four language pairs, chosen to cover both high-resource and lower-resource directions. While we also evaluate on unseen language pairs, broader multilingual training would be needed to determine how well the framework scales across a wider range of languages.

\paragraph{Heterogeneous human evaluation protocols.}
Our training data pools WMT years that differ in annotation protocol and score scale. We mitigate this by
normalising human targets within year--language-pair annotation cells and by reporting pairwise meta-evaluation
metrics on held-out WMT25 data. Nevertheless, protocol differences may still affect the learned combiner
through changes in annotation guidelines, domain composition, and rater behaviour.


\section*{Use of Generative AI}
AI was only used to assist in proofreading and revising manuscripts, and to provide code completions for experiments.

\bibliography{custom}
\appendix

\section{Related Work}\label{app:related-work}
\subsection{Single metric evaluation} 
Reference-based overlap metrics such as BLEU \citep{papineni-etal-2002-bleu} and chrF \citep{popovic-2015-chrf,popovic-2017-chrf} are commonly used in WMT Shared Tasks due to their efficiency and historical usage. However, their emphasis on lexical overlap means that legitimate semantic or syntactic variations can be under-rewarded, leading to poorer correlation with human evaluation \citep{freitag-etal-2022-results}.

Learned metrics, including BLEURT \citep{sellam-etal-2020-bleurt,pu-etal-2021-learning} and COMET \citep{rei-etal-2020-comet}, improve correlation with human judgments by using pretrained encoders and supervised calibration to human ratings. YiSi-1 \citep{lo-2019-yisi} provides an embedding-based similarity framework that can support both lower and higher-resource settings. More recently, neural generative models such as MetricX-23/24~\citep{kocmi-etal-2023-findings,freitag-etal-2022-results} became state-of-the-art in WMT22/23, showing stronger and more consistent results across many domains (though not all). However, no single metric dominates all domains. \citet{freitag-etal-2024-llms} showed that LLM-era MT outputs expose brittleness in some metrics and underscored the value of robust calibration to MQM/ESA.


\subsection{Static ensembles via meta metrics} 
MetaMetrics-MT (MM) \citep{anugraha-etal-2024-metametrics} learn static weights, typically per language pair, and optimises correlation with human scores, often outperforming individual metrics. However, weights are typically trained per-language and optimised against a single objective, leaving room to adapt within-language variation such as domain or segment properties. DMM addresses this limitation by conditioning on source-derived contexts.

\section{Data processing and joins}
\label{app:data}
\paragraph{Sources.} We use the WMT Metrics Shared Task test sets from 2021--2024~\citep{freitag-etal-2024-llms}. We focus on English to Czech (en–cs) and English to Chinese (en–zh), English to Japanese (en-ja), and English to Ukrainian (en-uk) to test performance variation on languages with different resource levels and topologies. For example, Czech is lower-resource, synthetic alphabetic while Chinese is higher resource, analytic, logographic language. For each language pair, we use the source sentences, system outputs and reference translations as input, and compare against the official segment-level human annotations, using the year-specific fields: 2021 \texttt{wmt-raw:seg}, 2022 \texttt{wmt-appraise:seg}, 2023 \texttt{da-sqm:seg}, 2024 \texttt{esa:seg}. Segments without human annotation are excluded.

\paragraph{Units.} A segment is one source sentence (WMT \texttt{seg-id}). A system is a MT run that produces one hypothesis per segment (\texttt{system-name}). Within a given (year, language pair) we identify an item by the pair (\texttt{system-name}, \texttt{seg-id}). When joining across years, we suffix the year to the identifier to keep years disjoint.

\section{Further details on soft-conditioned DMM}\label{app:soft-dmm}
We briefly recall that the model form over the $z$-standardised base-metric vector for instance $i$ is given by
$$
\hat y_i
=
\tilde{\mathbf{x}}_i^\top
\Bigl(
\mathbf{w}_0+\sum_{k=1}^{K} r_k(s_i;T)\mathbf{v}_k
\Bigr)
+b.
$$
Equivalently, this may be viewed as a linear model over an expanded design matrix consisting of the original metric features together with the interaction features
$$
\tilde{\mathbf{x}}_i\,r_1(s_i;T),\ldots,\tilde{\mathbf{x}}_i\,r_K(s_i;T).
$$

In this subsection, we review some theoretical and implementation details. 

\paragraph{Why a linear model?}
We use a \emph{linear} contextual model, rather than a more expressive nonlinear architecture, to isolate the effect of replacing discrete source conditioning with continuous source conditioning while preserving interpretability. Under this parameterisation, one may directly inspect the global baseline weights $\mathbf{w}_0$ and the context-specific perturbations $\mathbf{v}_k$, making it possible to ask how the preferred mixture of metrics changes across source contexts.

\paragraph{Hard conditioning is a limiting case.}
Hard conditioning is a limiting case of this model. As $T\to 0^+$, the responsibility vector $\mathbf{r}(s;T)$ approaches a one-hot assignment concentrated on the nearest centroid. In that regime, $\mathbf{w}_{\mathrm{eff}}(s;T)\to \mathbf{w}_0+\mathbf{v}_{c(s)},\ c(s)=\arg\min_k D_k(s),$ so prediction reduces to applying a single context-specific weight vector. Thus, the soft-conditioned model generalises the \emph{selection mechanism} of hard conditioning, while allowing intermediate cases in which a source segment partially belongs to several source contexts.

\section{Implementation details and hyperparameters}
\label{app:implementation}

This appendix specifies the implementation choices and hyperparameters required to reproduce the clustering, responsibility construction, and model training procedures described in Section \ref{sec:method}. 

\subsection{Source embeddings}
We embed each source segment using LaBSE sentence embeddings \citep{feng-etal-2022-language}. We encode source sentences in batches of 128 and apply embedding normalisation to unit length prior to clustering and distance computation. We treat the encoder as fixed and do not fine-tune it.

\subsection{Clustering via $k$-means on training sources}
For each choice of number of clusters $K$, we fit a $k$-means model to source embeddings from the training split only. We treat a source sentence as a unique string and deduplicate before fitting. We use Euclidean distance in embedding space. Since embeddings are normalised, Euclidean distance is proportional to angular distance, which improves stability for clustering and subsequent distance-based computations. For $K \in \{2,3,4,5,6,7\}$, we run $k$-means with the standard initialisation in \texttt{sklearn} \citep{scikit-learn}.


\subsection{Hard-conditioned models}
Hard-conditioned DMM realizes the four ensemble configurations described in Section~\ref{sec:hard_conditioned} by training separate models for each condition, without parameter sharing.
\begin{itemize}
  \item \textbf{Linear (OLS).} $M(x)=\mathbf{w}^{\top}\mathbf{s}(x)+b$. Inputs are z-standardized per metric and a linear regression model is fit using ordinary least squares to minimize mean squared error (MSE) on the human gold scores $h(x)$.
  \item \textbf{Neural (MLP).} Two hidden layers (64 and 32 units) with ReLU activations and dropout $p=0.2$. Trained with MSE using Adam (learning rate $10^{-3}$), batch size 32, for 100 epochs. Inputs are $z$-standardized.
  \item \textbf{Gaussian Process (GP).} A linear combiner $M(x)=\mathbf{w}^{\top}\mathbf{s}(x)$ with $\mathbf{w}\in[0,1]^d$. We learn $\mathbf{w}$ via Bayesian optimization with a Gaussian Process to \emph{maximize} Kendall's tau correlation to human scores. We use 5 random initial evaluations and 100 optimization iterations. Degenerate zero solutions are avoided with a small uniform initialization.
\end{itemize}

\subsection{Soft-conditioned DMM}

\paragraph{Optimisation.}
We fit the model by ridge regression with regularisation parameter $\alpha=1.0$. Only the base-metric features are standardised; the centroid-distance features are left on their original scale and are used solely to construct the responsibilities. The intercept is fit jointly with the regression coefficients. Our default implementation uses \texttt{sklearn}'s ridge regression solver.

\paragraph{Temperature selection.}
The temperature $T$ governs the sharpness of the responsibility distribution. We select $T$ on the validation split from the grid $$T\in\{0.1,\ 0.25,\,0.5,\ 1.0,\ 2.0\}.$$
For each candidate value of $T$, we:
\begin{enumerate}
    \item compute the corresponding responsibility vectors from the stored centroid distances,
    \item fit the ridge-regression contextual model on the training split,
    \item evaluate the fitted model on the validation split.
\end{enumerate}
We then retain the temperature yielding the best validation performance according to a specified selection criterion. In our experiments, we use validation Pearson correlation as the default selection criterion. 


\section{Full Table References}
\label{app:full_table_references}
\begin{table*}[!t]
    \centering
    \small
    \begin{tabular}{lcc cc cc cc cc cc}
    \toprule
    \multirow{2}{*}{\textbf{System}} & \multicolumn{2}{c}{\textbf{CS-DE}} & \multicolumn{2}{c}{\textbf{CS-UK}} & \multicolumn{2}{c}{\textbf{EN-AR}} & \multicolumn{2}{c}{\textbf{EN-ET}}  & \multicolumn{2}{c}{\textbf{EN-IS}}  & \multicolumn{2}{c}{\textbf{EN-IT}}\\
    \cmidrule(lr){2-3}\cmidrule(lr){4-5}\cmidrule(lr){6-7}\cmidrule(lr){8-9}\cmidrule(lr){10-11}\cmidrule(lr){12-13}
     & $\boldsymbol{acc^*_{eq}}$ & \textbf{SPA} & $\boldsymbol{acc^*_{eq}}$ & \textbf{SPA} & $\boldsymbol{acc^*_{eq}}$ & \textbf{SPA} & $\boldsymbol{acc^*_{eq}}$ & \textbf{SPA} & $\boldsymbol{acc^*_{eq}}$ & \textbf{SPA} & $\boldsymbol{acc^*_{eq}}$ & \textbf{SPA}  \\
    \midrule
    \multicolumn{13}{l}{\textsc{Meta-Metrics}} \\
    \quad GP & \cellcolor{blue!21}0.593 & \cellcolor{blue!30}\textbf{0.884} & \cellcolor{blue!23}0.551 & \cellcolor{blue!20}0.779 & 0.308 & 0.420 & \cellcolor{blue!18}0.645 & \cellcolor{blue!25}0.861 & \cellcolor{blue!18}0.751 & \cellcolor{blue!30}0.903 & \cellcolor{blue!1}0.574 & \cellcolor{blue!30}\textbf{0.787} \\
    \quad OLS & \cellcolor{red!17}0.449 & \cellcolor{red!21}0.258 & \cellcolor{red!30}0.461 & \cellcolor{red!30}0.372 & \cellcolor{red!21}0.174 & \cellcolor{red!27}0.150 & \cellcolor{red!18}0.497 & \cellcolor{red!12}0.511 & \cellcolor{red!11}0.548 & \cellcolor{red!11}0.538 & \cellcolor{red!30}0.564 & \cellcolor{red!4}0.746 \\
    \quad MLP & \cellcolor{blue!30}\textbf{0.602} & \cellcolor{blue!29}0.879 & \cellcolor{blue!30}\textbf{0.555} & \cellcolor{blue!12}0.751 & \cellcolor{blue!21}0.436 & \cellcolor{blue!20}0.705 & \cellcolor{blue!30}\textbf{0.659} & \cellcolor{blue!27}0.867 & \cellcolor{blue!29}0.770 & \cellcolor{blue!29}0.902 & \cellcolor{blue!30}\textbf{0.579} & 0.762 \\
    \midrule
    \multicolumn{13}{l}{\textsc{DMM $k=6$}} \\
    \quad GP & \cellcolor{blue!6}0.566 & \cellcolor{blue!19}0.835 & \cellcolor{blue!25}0.552 & \cellcolor{blue!22}0.787 & \cellcolor{red!4}0.231 & \cellcolor{red!3}0.300 & \cellcolor{blue!7}0.626 & \cellcolor{blue!7}0.793 & \cellcolor{blue!9}0.729 & \cellcolor{blue!23}0.885 & 0.571 & \cellcolor{red!2}0.749 \\
    \quad OLS & \cellcolor{red!30}0.434 & \cellcolor{red!30}0.219 & \cellcolor{red!30}0.461 & \cellcolor{red!30}0.373 & \cellcolor{red!30}0.158 & \cellcolor{red!30}0.141 & \cellcolor{red!30}0.483 & \cellcolor{red!30}0.454 & \cellcolor{red!30}0.512 & \cellcolor{red!30}0.478 & \cellcolor{red!1}0.569 & \cellcolor{red!9}0.742 \\
    \quad MLP & \cellcolor{blue!30}\textbf{0.602} & \cellcolor{blue!28}0.878 & \cellcolor{blue!23}0.551 & \cellcolor{blue!11}0.747 & \cellcolor{blue!30}\textbf{0.452} & \cellcolor{blue!30}\textbf{0.745} & \cellcolor{blue!29}0.658 & \cellcolor{blue!30}\textbf{0.874} & \cellcolor{blue!30}\textbf{0.771} & \cellcolor{blue!30}\textbf{0.904} & \cellcolor{blue!20}0.578 & \cellcolor{red!1}0.751 \\
    \quad CL & \cellcolor{blue!10}0.577 & \cellcolor{blue!14}0.812 & \cellcolor{blue!17}0.547 & \cellcolor{blue!30}\textbf{0.807} & 0.270 & \cellcolor{red!1}0.332 & \cellcolor{blue!10}0.631 & \cellcolor{blue!4}0.768 & \cellcolor{blue!8}0.726 & \cellcolor{blue!24}0.889 & \cellcolor{blue!1}0.574 & \cellcolor{red!30}0.733 \\
    \addlinespace[2pt]
    \midrule
    \multirow{2}{*}{\textbf{System}} & \multicolumn{2}{c}{\textbf{EN-KO}} & \multicolumn{2}{c}{\textbf{EN-MAS}} & \multicolumn{2}{c}{\textbf{EN-RU}} & \multicolumn{2}{c}{\textbf{EN-SR}}  & \multicolumn{2}{c}{\textbf{JA-ZH}}  & \multicolumn{2}{c}{\textbf{Avg}}\\
    \cmidrule(lr){2-3}\cmidrule(lr){4-5}\cmidrule(lr){6-7}\cmidrule(lr){8-9}\cmidrule(lr){10-11}\cmidrule(lr){12-13}
     & $\boldsymbol{acc^*_{eq}}$ & \textbf{SPA} & $\boldsymbol{acc^*_{eq}}$  & \textbf{SPA} & $\boldsymbol{acc^*_{eq}}$  & \textbf{SPA} & $\boldsymbol{acc^*_{eq}}$ & \textbf{SPA} & $\boldsymbol{acc^*_{eq}}$ & \textbf{SPA} & $\boldsymbol{acc^*_{eq}}$ & \textbf{SPA}  \\
    \midrule
    \multicolumn{13}{l}{\textsc{Meta-Metrics}} \\
    \quad GP & \cellcolor{blue!12}0.340 & \cellcolor{blue!26}0.872 & \cellcolor{red!24}0.295 & \cellcolor{blue!1}0.583 & \cellcolor{blue!30}\textbf{0.574} & \cellcolor{blue!30}\textbf{0.705} & \cellcolor{blue!30}\textbf{0.713} & \cellcolor{blue!30}\textbf{0.906} & \cellcolor{blue!6}0.456 & \cellcolor{blue!5}0.730 & \cellcolor{blue!11}0.536 & \cellcolor{blue!19}0.765 \\
    \quad OLS & \cellcolor{red!15}0.262 & \cellcolor{red!12}0.345 & 0.321 & \cellcolor{blue!30}\textbf{0.623} & \cellcolor{red!20}0.496 & \cellcolor{red!26}0.456 & \cellcolor{red!3}0.462 & \cellcolor{red!4}0.416 & \cellcolor{red!29}0.323 & \cellcolor{red!29}0.241 & \cellcolor{red!12}0.422 & \cellcolor{red!13}0.430 \\
    \quad MLP & \cellcolor{blue!30}\textbf{0.353} & \cellcolor{blue!30}\textbf{0.886} & \cellcolor{red!3}0.307 & \cellcolor{red!20}0.510 & \cellcolor{blue!19}0.568 & \cellcolor{blue!17}0.682 & \cellcolor{blue!29}0.712 & \cellcolor{blue!28}0.900 & \cellcolor{blue!30}\textbf{0.490} & \cellcolor{blue!30}\textbf{0.867} & \cellcolor{blue!29}0.558 & \cellcolor{blue!30}\textbf{0.794} \\
    \midrule
    \multicolumn{13}{l}{\textsc{DMM $k=6$}} \\
    \quad GP & \cellcolor{blue!6}0.332 & \cellcolor{blue!9}0.785 & \cellcolor{red!30}0.293 & \cellcolor{red!4}0.531 & \cellcolor{blue!12}0.563 & \cellcolor{blue!16}0.680 & \cellcolor{blue!7}0.648 & \cellcolor{blue!11}0.815 & \cellcolor{blue!6}0.456 & \cellcolor{blue!6}0.733 & \cellcolor{blue!2}0.513 & \cellcolor{blue!7}0.713 \\
    \quad OLS & \cellcolor{red!30}0.252 & \cellcolor{red!30}0.265 & \cellcolor{red!2}0.309 & \cellcolor{blue!7}0.600 & \cellcolor{red!30}0.491 & \cellcolor{red!30}0.450 & \cellcolor{red!30}0.375 & \cellcolor{red!30}0.251 & \cellcolor{red!30}0.322 & \cellcolor{red!30}0.239 & \cellcolor{red!30}0.402 & \cellcolor{red!30}0.381 \\
    \quad MLP & \cellcolor{blue!27}0.351 & \cellcolor{blue!29}0.883 & \cellcolor{red!1}0.311 & \cellcolor{red!30}0.502 & \cellcolor{blue!13}0.564 & \cellcolor{blue!9}0.663 & \cellcolor{blue!29}0.711 & \cellcolor{blue!28}0.898 & \cellcolor{blue!30}\textbf{0.490} & \cellcolor{blue!29}0.863 & \cellcolor{blue!30}\textbf{0.559} & \cellcolor{blue!30}\textbf{0.794} \\
    \quad CL & \cellcolor{blue!10}0.337 & \cellcolor{blue!14}0.818 & \cellcolor{blue!30}\textbf{0.346} & \cellcolor{blue!22}0.617 & \cellcolor{blue!10}0.561 & \cellcolor{blue!1}0.619 & \cellcolor{blue!19}0.690 & \cellcolor{blue!20}0.863 & \cellcolor{blue!8}0.460 & \cellcolor{blue!6}0.740 & \cellcolor{blue!7}0.528 & \cellcolor{blue!8}0.723 \\
    \bottomrule
    \end{tabular}
    \caption{System- and segment-level meta-evaluation results on unseen language pairs using soft pairwise accuracy (SPA) and group-by-item pairwise accuracy with tie calibration ($acc^*_{eq}$). All models are evaluated in a language-pooled setting. Higher values indicate better agreement with human judgements.}
    \label{tab:unseen_language_pairs}
\end{table*}
\begin{table*}[!t]
\centering
\footnotesize
\setlength{\tabcolsep}{3pt}
\renewcommand{\arraystretch}{1}
\begin{tabular}{l l r r p{2.8cm} p{3.5cm} p{3.2cm}}
\toprule
\textbf{$K$}
  & \textbf{Cl.}
  & \textbf{$|S|$}
  & \textbf{Med.\ Tok}
  & \textbf{Domains}
  & \textbf{GP Top}
  & \textbf{MLP Top} \\
\midrule
--- & \emph{Global} & 6{,}496 & --- & All
  & MetricX-24-L, -XL, YiSi-R
  & COMET \\
\midrule
2 & 0 & 3{,}686 & 12 & Social, e-comm., minutes
  & MetricX-24-L, XCOMET
  & XCOMET, chrF \\
  & 1 & 2{,}810 & 30 & Literary, news
  & MetricX-24-L, -XL, YiSi-R
  & COMET \\
\midrule
4 & 0 & 2{,}253 & \phantom{0}9 & E-comm., social, news
  & BLEU, MetricX-24-L
  & MetricX-24-L \\
  & 1 & \phantom{0}951 & 23 & News, literary
  & MetricX-24-L
  & MetricX-24-L \\
  & 2 & 1{,}799 & 17 & Social, minutes, e-comm.
  & XCOMET, COMET
  & spBLEU \\
  & 3 & 1{,}493 & 42 & Literary, news
  & COMET
  & COMET \\
\midrule
6 & 0 & 1{,}393 & 17 & Minutes, e-comm., social
  & MetricX-24-L, XCOMET
  & spBLEU \\
  & 1 & \phantom{0}871 & 11 & Social, news, e-comm.
  & BLEU, MetricX-24-L, -XL
  & MetricX-24-L, -XL \\
  & 2 & \phantom{0}998 & 25 & News (NYT, BBC)
  & MetricX-24-L
  & MetricX-24-L \\
  & 3 & \phantom{0}789 & 24 & News, literary
  & MetricX-24-L
  & MetricX-24-L \\
  & 4 & 1{,}482 & \phantom{0}7 & E-comm., social, minutes
  & BLEU
  & BLEU, spBLEU \\
  & 5 & \phantom{0}963 & 66 & Literary
  & COMET
  & COMET \\
\midrule
7 & 0 & \phantom{0}714 & 24 & UK/intl.\ news, literary
  & MetricX-24-L, -XL
  & chrF, BLEURT \\
  & 1 & 1{,}072 & \phantom{0}7 & Social, minutes, news
  & MetricX-24-L
  & MetricX-24-L \\
  & 2 & \phantom{0}847 & 15 & News, e-comm.
  & MetricX-24-L, -XL, YiSi-R
  & MetricX-24-L \\
  & 3 & 1{,}047 & 19 & Social, minutes
  & MetricX-24-XL, COMET
  & XCOMET, COMET \\
  & 4 & \phantom{0}773 & 19 & E-comm., minutes
  & MetricX-24-L, XCOMET, COMET
  & XCOMET, spBLEU, MetricX-24-L \\
  & 5 & \phantom{0}792 & \phantom{0}9 & Social, e-comm., minutes
  & MetricX-24-L
  & MetricX-24-XL \\
  & 6 & 1{,}251 & 45 & Literary, news
  & COMET
  & COMET \\
\bottomrule
\end{tabular}
\caption{
  Cluster statistics and top-ranked metrics across $K \in \{2,4,6,7\}$.
  $|S|$: unique source segments; Med.\ Tok: median token count.
  The Global row shows unclustered baseline weights.
  GP global weights are identical across all $K$;
  MLP global weights vary slightly across runs
  (COMET and spBLEU alternate as top-ranked within 0.04),
  so we report the modal ranking.
  Metrics are listed when within 0.01 of the top weight in each model.
}
\label{tab:cluster_metrics_appendix}
\end{table*}

\end{document}